\def\year{2022}\relax
\documentclass[letterpaper]{article} 
\usepackage{aaai22}  
\usepackage{times}  
\usepackage{helvet}  
\usepackage{courier}  
\usepackage[hyphens]{url}  
\usepackage{graphicx} 
\usepackage{natbib}  
\usepackage{caption} 
\DeclareCaptionStyle{ruled}{labelfont=normalfont,labelsep=colon,strut=off} 
\usepackage{algorithm}
\usepackage{algorithmic}
\usepackage{xcolor}

\usepackage{newfloat}
\usepackage{caption}
\usepackage{subcaption}
\usepackage{listings}
\usepackage{tabularx}

\newcommand\clearrow{\global\let\rowmac\relax}
\clearrow

\floatstyle{ruled}
\newfloat{listing}{tb}{lst}{}
\floatname{listing}{Listing}

\newcommand{\cy}[1]{}
\newcommand{\sk}[1]{}

\title{Using Sampling to Estimate and Improve Performance of\\ Automated Scoring Systems with Guarantees}
\author{
    Yaman Kumar Singla\textsuperscript{\rm 1,2,3}\equalcontrib,
    Sriram Krishna\textsuperscript{\rm 1}\equalcontrib,
    Rajiv Ratn Shah\textsuperscript{\rm 1},
    Changyou Chen\textsuperscript{\rm 3}
}
\affiliations{
    \textsuperscript{\rm 1}IIIT-Delhi, \textsuperscript{\rm 2}Adobe Media Data Science Research, \textsuperscript{\rm 3}State University of New York at Buffalo\\
    ykumar@adobe.com, sriramsk1999@gmail.com, rajivratn@iiitd.ac.in, changyou@buffalo.edu
}

\usepackage{multirow} 
\usepackage{booktabs} 

\begin{document}
\maketitle

\begin{abstract}
Automated Scoring (AS), the natural language processing task of scoring essays and speeches in an educational testing setting, is growing in popularity and being deployed across contexts from government examinations to companies providing language proficiency services. However, existing systems either forgo human raters entirely, thus harming the reliability of the test, or score every response by both human and machine thereby increasing costs. We target the spectrum of possible solutions in between, making use of both humans and machines to provide a higher quality test while keeping costs reasonable to democratize access to AS. In this work, we propose a combination of the existing paradigms, sampling responses to be scored by humans intelligently. We propose reward sampling and observe significant gains in accuracy (19.80\% increase on average) and quadratic weighted kappa (QWK) (25.60\% on average) with a relatively small human budget (30\% samples) using our proposed sampling. The accuracy increase observed using standard random and importance sampling baselines are 8.6\% and 12.2\% respectively. Furthermore, we demonstrate the system's model agnostic nature by measuring its performance on a variety of models currently deployed in an AS setting as well as pseudo models. Finally, we propose an algorithm to estimate the accuracy/QWK with statistical guarantees\footnote{Our code is available at \url{https://git.io/J1IOy}}.
\end{abstract}

\section{Introduction}

\begin{figure}[t]
\centering
\includegraphics[width=0.9\columnwidth]{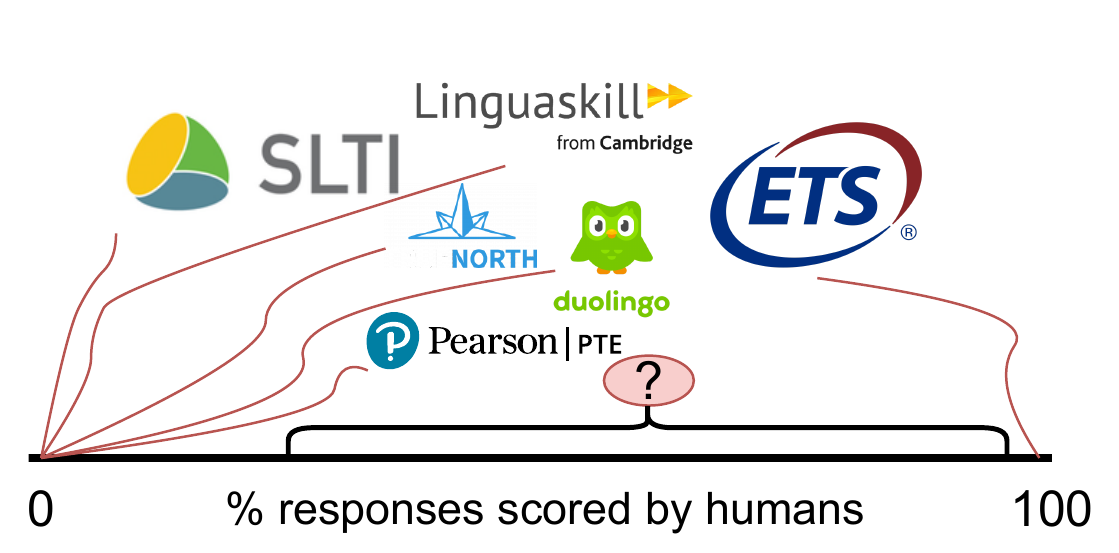}
\caption{Existing Automated Scoring systems\protect \footnotemark~either do not involve humans at all in their scoring (Duolingo, Second Language Testing Inc (SLTI)), or utilize human raters for every single response (Educational Testing Services (ETS)). Crucially, there are no solutions that target the gulf in between, where humans are involved in scoring only some percentage of the responses.}
\label{fig:types_of_as}
\end{figure}
\footnotetext{TOEFL by ETS, Pearson PTE, SLTI, Linguaskill by Cambridge, Duolingo English Test, and TrueNorth by Emmersion are registered brand names and are shown here for illustration purposes only. The authors claim no rights over their logos or brand names. In this work, we mainly refer to the automatically scored speaking and writing proficiency measurement tests of these companies.}

Automated Scoring (AS), the task of assigning scores to unstructured responses to open-ended questions, is an NLP application typically deployed in an educational setting. Historically, its origins have been traced to the work of Ellis Page \cite{page1967statistical}, who first argued for the possibility of scoring essays by computer. The factors behind the rise of Automated Scoring systems and its subtasks, Automated Essay Scoring (AES) and Automated Speech Scoring (ASS) are numerous, including but not limited to, the costs involved in providing and scoring a test, and ensuring that all test takers are scored on a uniform set of rubrics applied across all students, standardizing the scoring for these unstructured responses. The promise of lower costs and uniform scoring rubrics among other factors, has fueled the popularity of Automated Scoring systems, and various ML and DL systems are being increasingly deployed in AS contexts \cite{kumar2019get,liu2019automated,singla2021speaker}. AS systems are behind some of the world's most popular language tests, such as ETS' Test of English as a Foreign Language (TOEFL) \cite{zechner2009automatic}, Duolingo's English Test (DET) \cite{laflair2019duolingo}, among others. Various governmental institutions and businesses have also instituted automated systems to augment the scoring process, such as the state schools of Utah \cite{utahcompose} and Ohio \cite{odonnell2018ohio}, and a majority of BPOs. It is estimated that automatic scoring has a large market size of more than USD 110 billion, with a US market size alone of USD 17.1 billion \cite{researchAndMarketsEdTesting,causeIQETSWorth,educationTestingSalaries,ibisWorldEdTesting}.

\begin{figure*}[t]
    \centering
    \includegraphics[width=\textwidth]{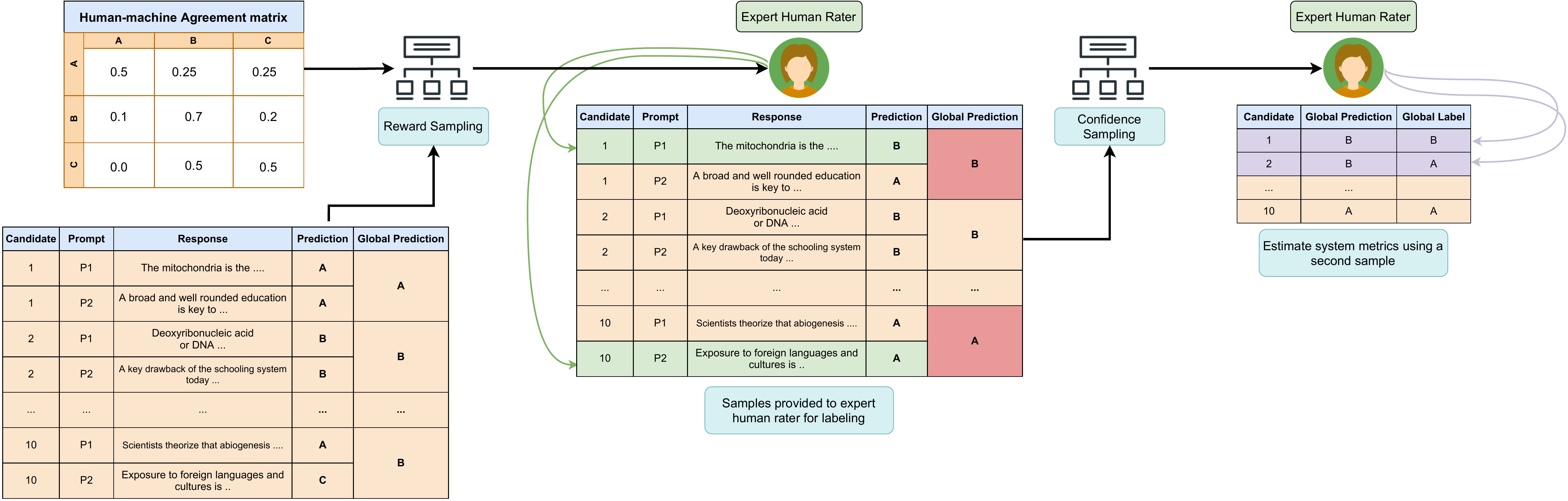}
    \caption{From a dataset, records are sampled and assigned to expert human raters for double scoring based on a \textit{human-machine agreement matrix}. A second sample is then drawn to check predictions and metrics are estimated with statistical guarantees.}
    \label{system_diagram}
\end{figure*}

However, this popularity has not been without backlash, with criticism focusing on different aspects, such as ``the overreliance on surface features of responses, the insensitivity to the content of responses and to creativity, and the vulnerability to new types of cheating and test-taking strategies.'' \cite{yang2002review}. Others have given harsher criticisms, such as \cite{perelman}, who shows that it is possible to \textit{game} the system and achieve near perfect scores on ETS and Vantage Technologies' AES systems with gibberish prose. This has led to the revoking of NAPLAN AES in Australia \cite{naplan}.

Nonetheless, the ability of AS systems to instantly provide scores, reduce costs, and make language proficiency tests more widely available to all, makes them an important research area and subsequently there is considerable interest in improving them across multiple dimensions, from leveraging advancements in NLP to achieve state-of-the-art performance \cite{liu2019automated} to improving their robustness \cite{kumar2020calling,parekh2020my,singla2021aes}. In this work, we tackle another facet of Automatic Scoring systems, that of improving performance by bringing humans into the loop.

Typically in an AS task, a test taker's responses are scored on prompts of varying difficulty levels. Each prompt has its own difficulty level, and based on the prompts' difficulty and the quality of the candidate's answers to these prompts, a score is assigned to the candidate. The Central European Framework of Reference for Languages (CEFR) is an international standard for measuring language proficiency and assigns scores on a six-point scale from \textbf{A1} (beginner) to \textbf{C2} (proficient), each score with their own rubrics for evaluation \cite{broeder2008language}. Each prompt and response is assigned a score on this scale and a \textbf{global} score is computed aggregating these individual scores.

\noindent Existing AS systems are typically of two varieties (Fig~\ref{fig:types_of_as}):\\
\indent \textbf{Double Scoring:} Examinations such as ETS' TOEFL score every response by one human and an AES system as the second rater. A second human rater resolves any disagreements between the two \cite{yang2002review}. This effectively means that atleast one human rater is required for every test, driving up costs, as evidenced by the TOEFL's high price of $\sim$230 USD \cite{etsTOEFLFee}.\\
\indent \textbf{Machine-only Scoring:} On the other end of the scale are tests like the Duolingo English Test (DET) which are scored by machines alone, without any human intervention, keeping costs low but decreasing the reliability of the test. This is one of the main reasons, the DET costs USD 49, less than one-fourth of what TOEFL costs. All tests surveyed in Fig~\ref{fig:types_of_as} except Pearson PTE are priced around the same price point.

Our solution (Fig~\ref{system_diagram}) proposes to unify these varieties, allocating the available human budget intelligently to balance the reliability of the test with the cost to the test-taker. To the best of our knowledge, no existing systems target this continuum of utilizing both humans and AS raters. Providing this option would allow AS models to be deployed in more versatile scenarios, working in tandem with expert human raters to provide both \textit{reliability} and \textit{lower-cost} solutions. Increasing reliability helps to build trust in automatically scored exams, thus leading to broader adoption. Cost is a critical consideration to lower-income test-takers and those who need to take the test multiple times.

We define the problem and solution more formally as follows: given a set of responses to be scored, a target AS model, and an \textbf{expert human budget} (that is, the number of responses we can have scored by expert human raters), our goal is to \textit{efficiently} sample responses to be scored by the expert. These expert-scored samples are then combined with automatically scored samples to \textit{maximize} the overall system performance metric. We propose a novel Monte-Carlo sampling based reward sampling algorithm to efficiently sample responses to maximize the system performance.

Usually one or multiple amongst accuracy, Quadratic Weighted Kappa (QWK), or Cohen's kappa \cite{taghipour2016neural, zhao2017memory, kumar2019get, grover2020multi,singla2021speaker} are used in automatic scoring literature as they are robust measures of inter-rater reliability, a primary goal in Automated Scoring. A key point to be noted is that the reliability of the test ({\it i.e.} how consistently a test measures a characteristic) is measured on the \textbf{global} score (the aggregate of the responses) and lesser on the score on the individual responses. The global score determines admissions, interviews, and career growth, while per-item scores are used as indicators of particular skills. While intuitively, we can say that there exists a monotonically increasing relationship between the reliability of the test on individual questions and the overall score, we show that it is more efficient to consider the global context instead of item-level context, while sampling responses for getting them double-scored by humans. 

We establish strong baselines using \textbf{Uncertainty Sampling} (\S\ref{sec:Uncertainty Sampling}), an importance sampling formulation that samples using probability of being wrong output by the AS model. We propose \textbf{Reward Sampling} (\S\ref{sec:Reward Sampling}), that samples based on the estimated \textit{reward} of correcting a mistake.


\noindent We summarize our main contributions as follows:

- We propose to combine existing paradigms to integrate humans with Automated Scoring systems. Provided a budget indicating the number of responses that can be scored by human raters, we observe significant gains in accuracy and QWK using our proposed sampling model, \textbf{Reward Sampling} (\S\ref{sec:Reward Sampling}). For instance, by using 40\% human budget with an AS model with 64\% accuracy, our sampling methodology can achieve an accuracy gain of 23\% while random sampling leads to 14\% and uncertainty sampling leads to 15\%. To the best of our knowledge, this is the first time such a formulation has been considered in Automatic Scoring systems.
    
- We conduct experiments on various models differing in accuracy to show our algorithm's model agnostic nature (\S\ref{sec:System Overview}). We include results from models deployed in AS settings in the real world to crafted pseudo models. Averaging over these models, we observe 19.80\% increase in accuracy and 25.60\% increase in QWK when using reward sampling with 30\% of the dataset as a human budget. The random sampling and uncertainty sampling baselines achieve 8.6\% and 12.2\% gains in accuracy, respectively.

- While augmenting the system's performance is an important goal, it is equally important to quantify this improvement, especially when deployed in the real world, where there are no labeled datasets to compare against and the consequences of misgrading, for both business and test takers, could be catastrophic. Thus, we also propose an algorithm to \textit{estimate} the accuracy and QWK achieved, with statistical guarantees. (\S\ref{sec:Estimation with Guarantees}). 

\section{Related Work}
\label{sec:related work}
Broadly, our paper covers two areas of research: Automatic Scoring and Sampling methods. Here we cover them briefly.
\paragraph{Automatic Scoring:} The
goal of an automatic scorer is to assess language competence of a candidate
with an accuracy matching that of a human grader, but faster, with greater consistency and at
a fraction of the cost \cite{malinin2019uncertainty,yan2020handbook}. Almost all work in the automatic scoring domain has been to better model the scoring of essays and speech traits as a natural language processing task. The techniques have ranged from manually-engineered natural language features \cite{kumar2019get,dong2016automatic} to LSTMs, memory networks \cite{zhao2017memory} and transformers \cite{singla2021speaker}. There has also been some recent work in other facets of AS including adversarial testing \cite{ding2020don,kumar2020calling,parekh2020my}, explainability \cite{kumar2020explainable}, uncertainty estimation \cite{malinin2019uncertainty}, off-topic detection \cite{malinin2016off}, evaluation metrics \cite{loukina2020using}, \textit{etc.} To the best of our knowledge, there is no work on increasing the reliability of automatic scoring systems by bringing humans into the loop. Most white papers from second language testing firms mention results on historical data as a measure of their reliability \cite{brenzel2017duolingo,pearsonWhitepaper}. However, historical results are not a guarantee for performance over time. Due to continuous domain shift, historical results cannot be trusted for a model's future performance gains. Therefore, performance guarantees of AS models are essential to establish institutional trust in them. To fill this research gap, we propose reward sampling based on Monte Carlo sampling methods for measuring and increasing AS systems' reliability.

\paragraph{Monte-Carlo Sampling For Evaluation:} There has been much work in improving automatic metrics using Monte-Carlo sampling methods in machine translation and natural language (NL) evaluation \cite{chaganty2018price,hashimoto2019unifying,wei2021statistical}. They use statistical sampling methods like importance sampling and control variates to combine automatic NL evaluation with expensive human queries. To the best of our knowledge, we are the first to extend sampling techniques in the context of automatic scoring. We use them to combine relatively cheaper automatic scoring model results with expensive human expert scorers. \citet{kang2020approximate} use sampling for approximate selection queries. They combine cheap classifiers with expensive estimators to meet minimum precision or recall targets with guarantees. We extend their work to take the global context into account while estimating accuracy (\S\ref{sec:Estimation with Guarantees}).

\section{System Overview}
\label{sec:System Overview}
This section describes the components of the proposed solution, the intuition and reasoning behind the sampling mechanisms, and the algorithm for estimating the metrics with statistical guarantees. Given an Automated Scoring model, a dataset to be scored, and a human budget indicating the percentage of records we can provide to expert human raters for scoring, records are sampled making use of a pre-computed \textbf{human-machine agreement matrix} (to be described below). For the samples selected, we replace the predictions made by the AS model with the scores provided by the human raters and compute an estimate of the increase in accuracy and QWK with guarantees (Fig~\ref{system_diagram}).

When considering sampling, the baseline approach is random sampling {\it i.e.} sampling with uniform probability for each record in the dataset. This is not a good allocation of resources, as when considering models of high quality, most samples will not provide any value. For example, with a model of 75\% accuracy, random sampling would only provide value for $\sim$25\% of samples, as the rest would have been correctly scored anyway. This motivates our search for a more efficient sampling mechanism, one that takes into account the probability of the model being wrong with respect to a human expert, and crucially, the reward that would be gained by correcting this mistake. We define the reward as the magnitude of the change in the \textit{global} score that would occur when a \textit{local} response is changed as a result of human correction of machine score (\S\ref{sec:Reward Sampling}).

\subsection{Human-Machine Agreement Matrix}
\label{sec:Human-Machine Agreement Matrix}

\begin{figure}[t]
\centering
\includegraphics[width=0.80\columnwidth]{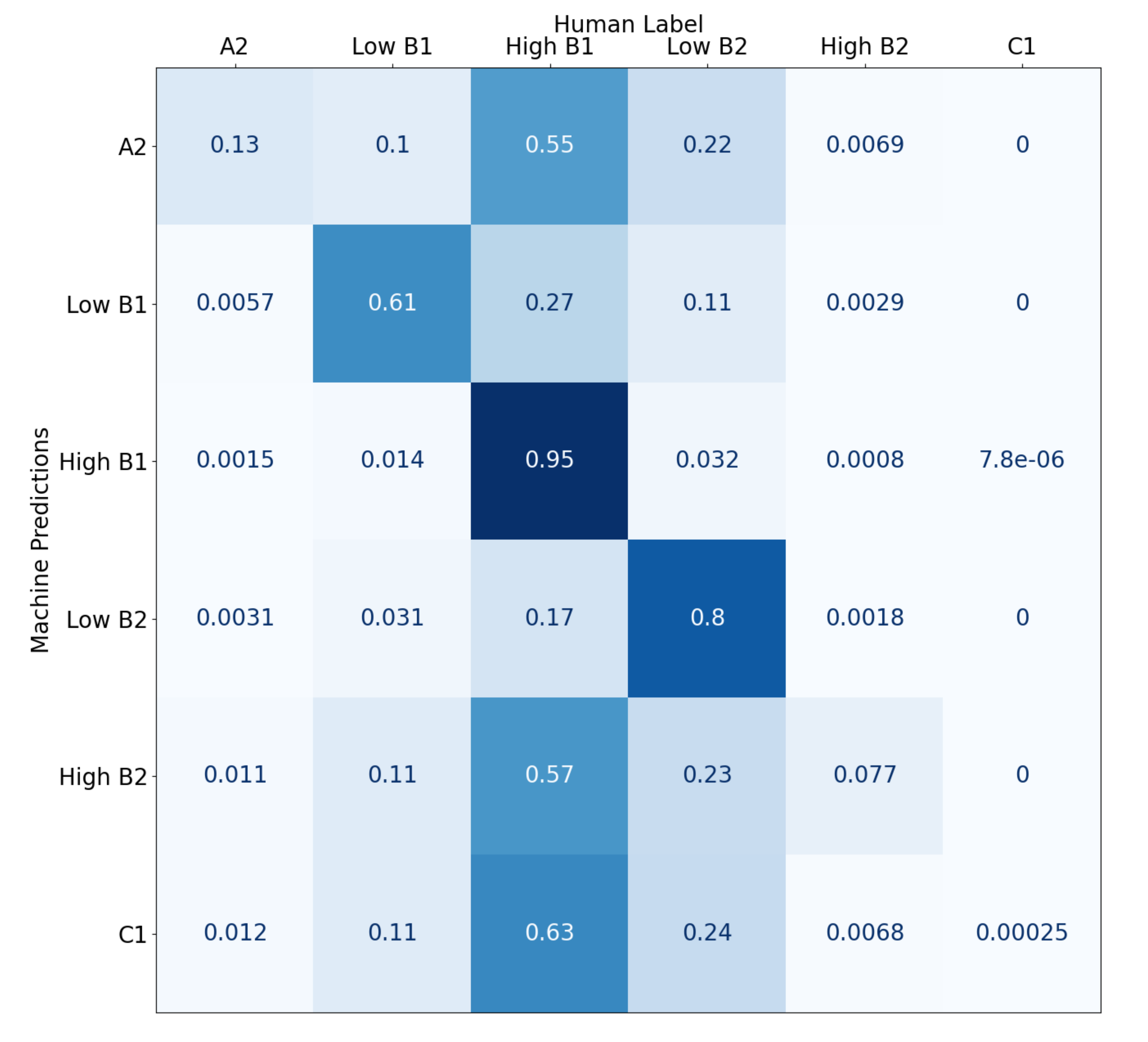}
\caption{A sample human-machine agreement matrix on a CEFR aligned scoring scale. The rows indicate machine predictions, and each row is normalized to give the probability of the machine class matching the human labeled class.}
\label{fig:human_machine_matrix}
\end{figure}

The human-machine agreement matrix is a normalized confusion matrix of the AS model's predictions and the ground truth, precomputed on validation data or historical test data. As the matrix is normalized, each entry indicates the probability of the class predicted by the machine aligning with the class labeled by the human. Fig~\ref{fig:human_machine_matrix} shows a sample human-machine agreement matrix where \texttt{m[Low B1][High B1] = 0.27} indicates the probability of the ground truth being \texttt{High B1} when the machine has predicted \texttt{Low B1}.

\subsection{Uncertainty Sampling}
\label{sec:Uncertainty Sampling}
The key idea behind uncertainty sampling is that the machine is not equally likely to be wrong across all prediction classes. Some scores may be assigned with much better accuracy than others. This idea is borne out by the human-machine agreement matrix as well, where the probabilities of a correct prediction are along the principal diagonal. We can see in Fig~\ref{fig:human_machine_matrix}, \texttt{High B1, Low B2} are accurately predicted whereas \texttt{A2, High B2} predictions are likely to be wrong. Since the machine is likely making a wrong judgement when predicting these classes, it would be more efficient to sample more from the records where these predictions have been made and corrected using human labelers.

To quantify this, we formulate Uncertainty Sampling as vanilla importance sampling, where the \textit{uncertainty} of each class is calculated using the cross-entropy function. Each row in the human-machine agreement matrix represents the probability distribution of the ground truth when that particular class has been predicted. The cross entropy of this distribution with the \textit{ideal} distribution (one-hot encoding for that class) is calculated. 

For \textit{e.g.}, the distribution associated with \texttt{Low B1} in the matrix is $[0.0057, 0.61, 0.27, 0.11, 0.0029, 0]$. The cross entropy of this distribution with the ideal distribution ( $[0, 1, 0, 0, 0, 0]$) for \texttt{Low B1} is calculated. In this way, we can quantify the ``loss'' associated with \texttt{Low B1}. Subsequently, every record is assigned a loss associated with the prediction made for that record, and this is normalized over the entire dataset to create a probability distribution. We draw a sample $s\sim U(D)$ without replacement from the uncertainty distribution over the dataset $U(D)$. The provided human budget indicates the number of samples to be drawn and the likelihood of a record being drawn corresponds to the uncertainty associated with the prediction class.

\begin{table*}
\centering
\caption{Accuracy (\textbf{acc}) and Quadratic Weighted Kappa (\textbf{kappa}) for various models across multiple sampling methods and increasing percentages of the dataset as sample size. \textbf{Bold} indicates the best performing variant for each configuration.}

\resizebox{0.9\textwidth}{!}{\begin{tabular}{c|c|c|c|c|c|c|c|c|c|c|c}
\toprule
\multirow{2}{10em}{Model metrics (\textbf{global})} & \multirow{2}{8em}{Sampling Method} & \multicolumn{5}{|c|}{Accuracy Improvement} & \multicolumn{5}{|c}{Kappa Improvement} \\
\multirow{2}{10em}{} & \multirow{2}{8em}{} & \multicolumn{5}{|c|}{Human Budget \textrightarrow} & \multicolumn{5}{|c}{Human Budget \textrightarrow} \\
& & 10\% & 20\% & 40\% & 60\% & 80\% & 10\% & 20\% & 40\% & 60\% & 80\% \\
\midrule
\multirow{3}{10em}{\textbf{BERT-Baseline} \\acc - 0.66; kappa - 0.56} & Random & 0.69 & 0.72 & 0.78 & 0.84 & 0.93 & 0.59 & 0.63 & 0.7 & 0.78 & 0.9 \\
& Uncertainty & 0.7 & 0.73 & 0.8 & 0.86 & 0.93 & 0.59 & 0.62 & 0.71 & 0.8 & 0.9 \\
& Reward & \textbf{0.74} & \textbf{0.82} & \textbf{0.88} & \textbf{0.91} & \textbf{0.95} & \textbf{0.64} & \textbf{0.76} & \textbf{0.84} & \textbf{0.88} & \textbf{0.94} \\
\hline
\multirow{3}{10em}{\textbf{BERT-Two Stage} \\acc - 0.69; kappa - 0.60} & Random & 0.73 & 0.75 & 0.81 & 0.87 & 0.93 & 0.65 & 0.68 & 0.75 & 0.83 & 0.91 \\
& Uncertainty & 0.72 & 0.75 & 0.82 & 0.87 & 0.94 & 0.63 & 0.66 & 0.74 & 0.82 & 0.92 \\
& Reward & \textbf{0.79} & \textbf{0.86} & \textbf{0.91} & \textbf{0.92} & \textbf{0.96} & \textbf{0.72} & \textbf{0.81} & \textbf{0.87} &\textbf{0.9} & \textbf{0.94} \\
\hline
\multirow{3}{10em}{\textbf{LSTM-Attn-Baseline} \\acc - 0.64; kappa - 0.54} & Random & 0.67 & 0.71 & 0.78 & 0.85 & 0.93 & 0.58 & 0.63 & 0.71 & 0.79 & 0.9 \\
& Uncertainty & 0.68 & 0.72 & 0.79 & 0.88 & 0.93 & 0.57 & 0.6 & 0.69 & 0.82 & 0.9 \\
& Reward & \textbf{0.73} & \textbf{0.78} & \textbf{0.87} & \textbf{0.92} & \textbf{0.96} &\textbf{ 0.62} & \textbf{0.71} & \textbf{0.83} & \textbf{0.89} & \textbf{0.95} \\
\hline
\multirow{3}{10em}{\textbf{LSTM-Attn-Two Stage} \\acc - 0.65; kappa - 0.57} & Random & 0.67 & 0.71 & 0.76 & 0.85 & 0.92 & 0.59 & 0.64 & 0.7 & 0.8 & 0.89 \\
& Uncertainty & 0.68 & 0.73 & 0.8 & 0.87 & 0.93 & 0.58 & 0.62 & 0.71 & 0.81 & 0.91 \\
& Reward & \textbf{0.74} & \textbf{0.82} & \textbf{0.87} & \textbf{0.9} & \textbf{0.95} & \textbf{0.66} & \textbf{0.75} & \textbf{0.83} & \textbf{0.86} & \textbf{0.93} \\
\hline
\multirow{3}{10em}{\textbf{Pseudo Model-0.75} \\acc - 0.72; kappa - 0.57} & Random & 0.74 & 0.76 & 0.8 & 0.86 & 0.93 & 0.62 & 0.64 & 0.72 & 0.81 & 0.9 \\
& Uncertainty & \textbf{0.82} & \textbf{0.9} & \textbf{0.93} &\textbf{ 0.97} & \textbf{0.98} & \textbf{0.73} & \textbf{0.85} & \textbf{0.9} & \textbf{0.95} & \textbf{0.98} \\
& Reward & \textbf{0.81} & 0.86 & \textbf{0.92} & \textbf{0.96} & \textbf{0.98} & \textbf{0.73} & 0.78 &\textbf{ 0.89} & \textbf{0.94} & \textbf{0.97} \\
\bottomrule
\end{tabular}}
\label{result_improving_metrics}
\end{table*}

\subsection{Reward Sampling}
\label{sec:Reward Sampling}

For single skill testing exams (for \textit{e.g.}, one out of speaking, writing, reading) like the one by \citet{slti} and \citet{lti}, the test reliability and validity are measured over the complete test as opposed to individual prompts. While increasing accuracy on individual prompts (through sampling and subsequent human intervention) is a sure way of increasing the accuracy on the overall exam, it is more efficient to directly sample records which are \textit{more likely to affect the overall result}, rather than simply sampling those which the machine is uncertain about. In uncertainty sampling, we sample records based on the likelihood of the prediction being wrong, but we do not consider whether \textit{being right} would actually change the global score. This is the motivation behind reward sampling. Here we sample records which are more likely to generate a larger reward, {\it i.e.}, a change in the score at the global level. To this end, the expected reward $E_R$ is calculated for each record in the dataset as:
    \begin{equation}
    E_R(d) = \sum_{c \in C} p(c \, | \, m)*reward(d, c) \;\; \forall d \in D
    \end{equation}
where $d$ represents one record in the dataset $D$, $c$ and $m$ represent classes in the set of all classes $C, \; p(c \, | \, m)$ indicates the probability of the ground truth being $c$ when machine has predicted $m$, and the reward function is denoted as $reward$. The expected reward encodes the reward gained by the ground truth being $c$ when the machine has predicted $m$ weighted by the \textit{probability} of the same, summed over every class $c$. $p(c \, | \, m)$ is looked up from the human-machine agreement matrix and the output of the reward function is weighted by this probability.

The reward function calculates the reward gained by swapping the predicted class with a different class. The aggregate label for the candidate associated with $d$ is calculated before and after the swap with a new class, and the reward is defined as the \textbf{absolute difference} between the two scores, which encodes the magnitude of the score change that would happen if the prediction class was changed from $m$ to $c$. The absolute difference is considered because it is equally important if the new score is greater or lesser than the predicted score, thus incurring the same reward. If the prediction is an outlier compared to predictions on other responses of the same candidate, a large reward could be generated when changing predictions, making it a prime target for sampling. On the other hand, if changing the class to $c$ does not change the final score, then a reward of $0$ would be generated. With a zero reward, these records would not be sampled. Thus, to ensure that every record has a nonzero reward \textit{i.e} a nonzero probability to be sampled, the reward is additively smoothed $E_R(d) = E_R(d) + \Delta$ where $\Delta = 0.001$. In this manner, an expected reward is calculated for each record in the dataset. 

The sampling procedure proceeds similarly: the rewards are normalized to create a probability distribution over which a sample $s \sim E_R(D)$ is drawn. In using this sampling mechanism, we directly sample records that are most likely to provide us an improvement at the aggregate level, compared to indirectly improving the aggregate metrics when using uncertainty sampling.  

\subsection{Estimation with Guarantees}
\label{sec:Estimation with Guarantees}

In high stakes testing scenarios, it is critical to ensure that the system does not fail catastrophically. For this reason, it is important to provide estimations of system metrics with guarantees. \citet{kang2020approximate} describes an algorithm that provides statistical guarantees on precision/recall on a subset of results returned from a dataset. More specifically, given a dataset, a precision/recall target value, sample size and a failure probability, the algorithm returns a result (a subset of the dataset) which meet the required precision/recall target with a \textit{probabilistic guarantee}. 

Our task is similar, but instead looks at providing guarantees on the accuracy/QWK of overall score on the entire dataset rather than just a dataset subset and individual samples. To provide these guarantees, we form confidence intervals over accuracy/QWK and take the lower bound. 

The samples selected by reward and uncertainty sampling procedures are not a good fit for estimation as they have been taken with the purpose of correcting mistakes and improving reliability. This means that highly underconfident samples would be selected, thus leading to inaccurate performance estimates. \citet{kang2020approximate} show that importance sampling based on a model's confidence of prediction improve over uniform random sampling by providing a lower variance estimate. More specifically, they show that the squared confidence of the model minimizes the variance of the estimate. As we have not considered model confidence in our work, we take the following formulation as a proxy for confidence, applied over every \textit{candidate} who wrote the test (not responses to individual questions):
\begin{equation}
\label{eq:estimation with guarantees}
    \zeta(t) = (1 - \sum_{i \in t} i[u]\,)^2 \;\; \forall t \in T 
\end{equation}
where $\zeta$ represents the confidence associated with a test taker $t$ in the set of all test-takers $T$, $i$ represents individual responses of $t$ and $u$ represents the uncertainty. From uncertainty sampling, we have a normalized uncertainty associated with each response, this is aggregated over all responses of a candidate, subtracted from 1 and then squared to provide a confidence estimate. This confidence is normalized to create a probability distribution. A secondary smaller sample is taken over this distribution of \textit{candidates}, effectively sampling all underlying responses of the candidate. Using the aggregated labels and predictions, the lower bound estimates of accuracy and kappa \cite{mchugh2012interrater} are calculated.

\begin{figure*}[!ht]
    \centering
    \begin{subfigure}[b]{0.4\textwidth}
    \centering
    \includegraphics[width=\columnwidth]{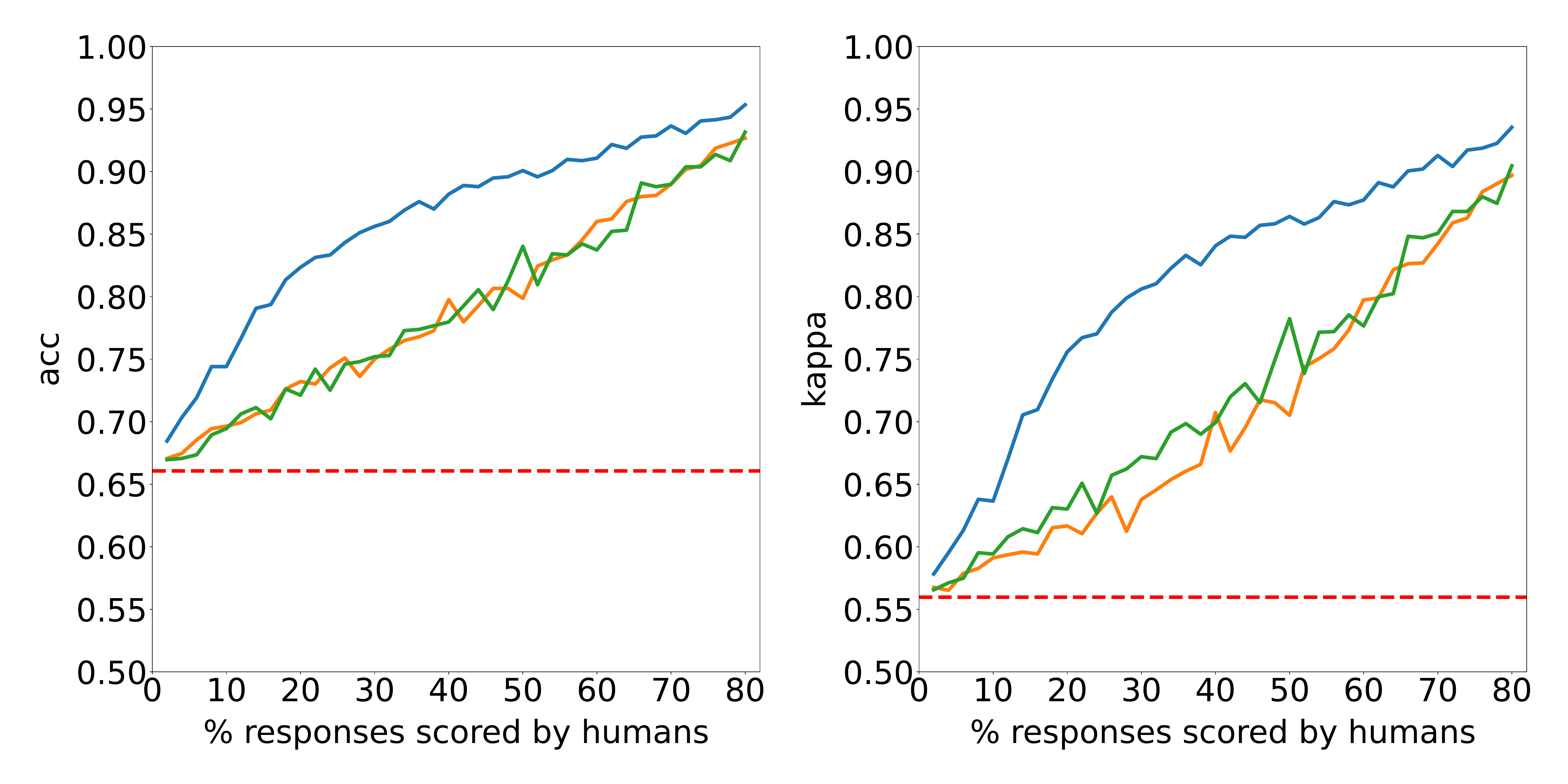}
    \caption{BERT-Baseline Model}
    \end{subfigure}%
    \begin{subfigure}[b]{0.4\textwidth}
    \centering
    \includegraphics[width=\columnwidth]{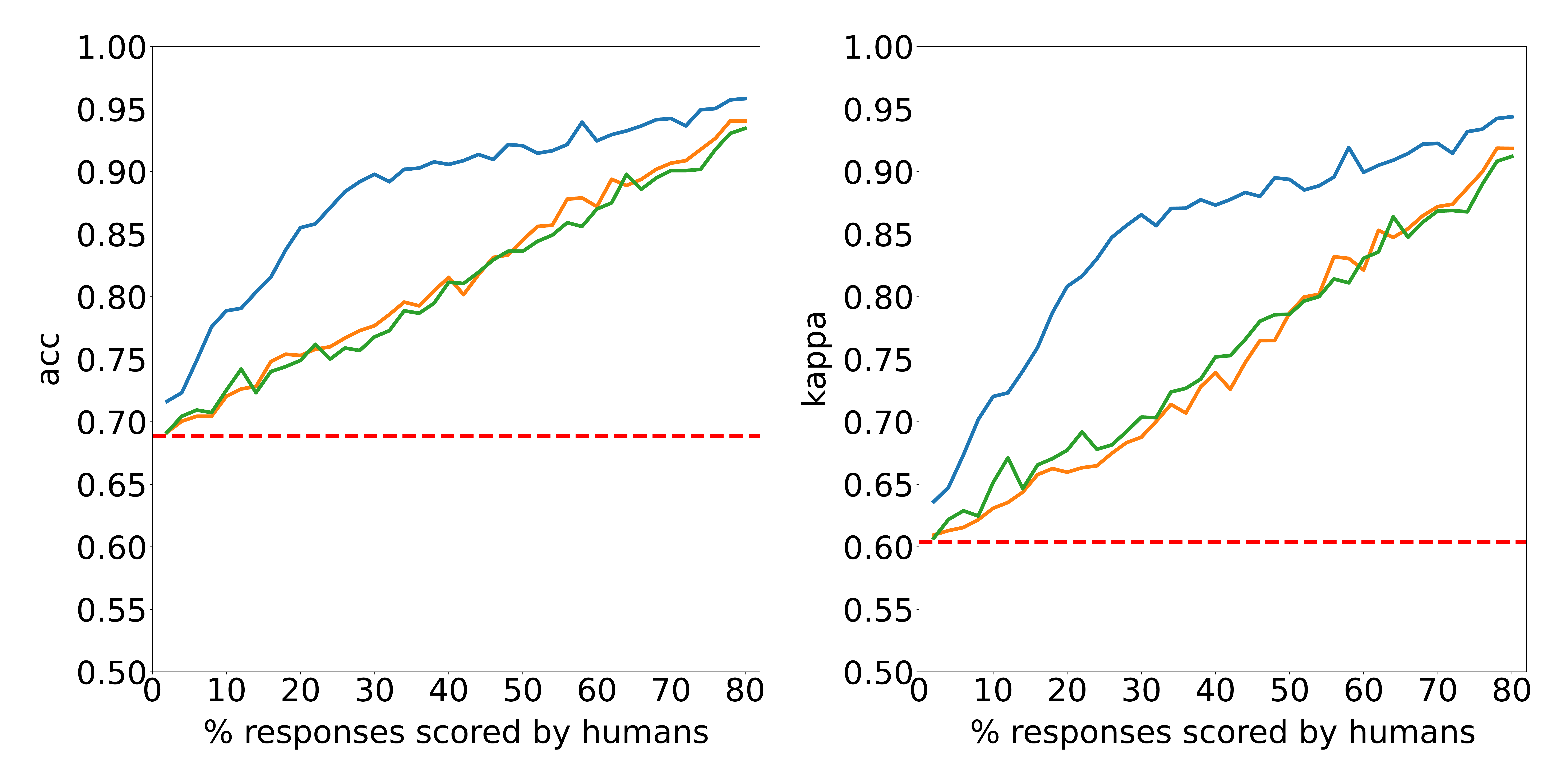}
    \caption{BERT-Two Stage Model}
    \end{subfigure}
    \begin{subfigure}[b]{0.4\textwidth}
    \includegraphics[width=\columnwidth]{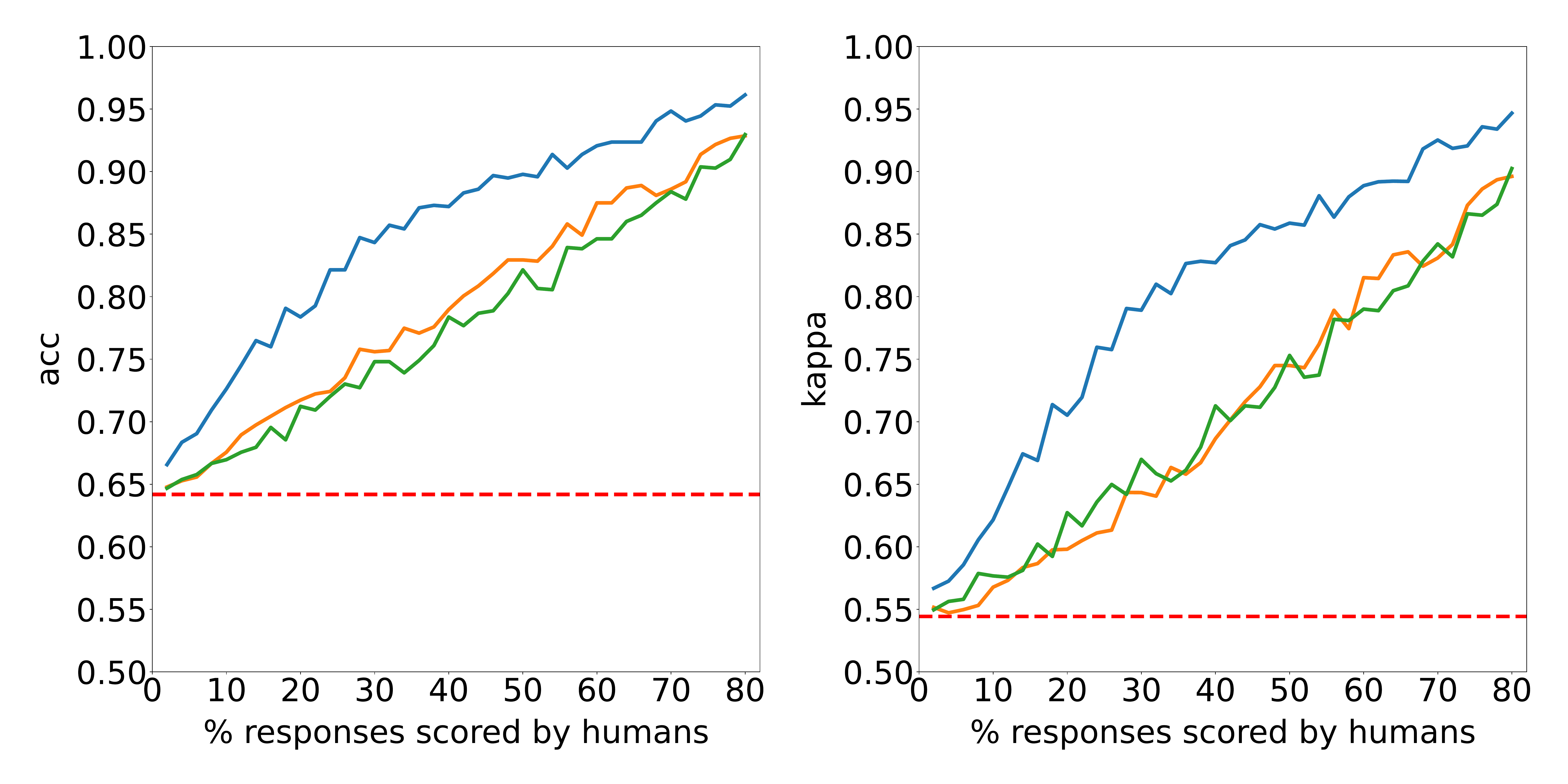}
    \caption{LSTM-Attention-Baseline Model}
    \end{subfigure}%
    \begin{subfigure}[b]{0.4\textwidth}
    \includegraphics[width=\columnwidth]{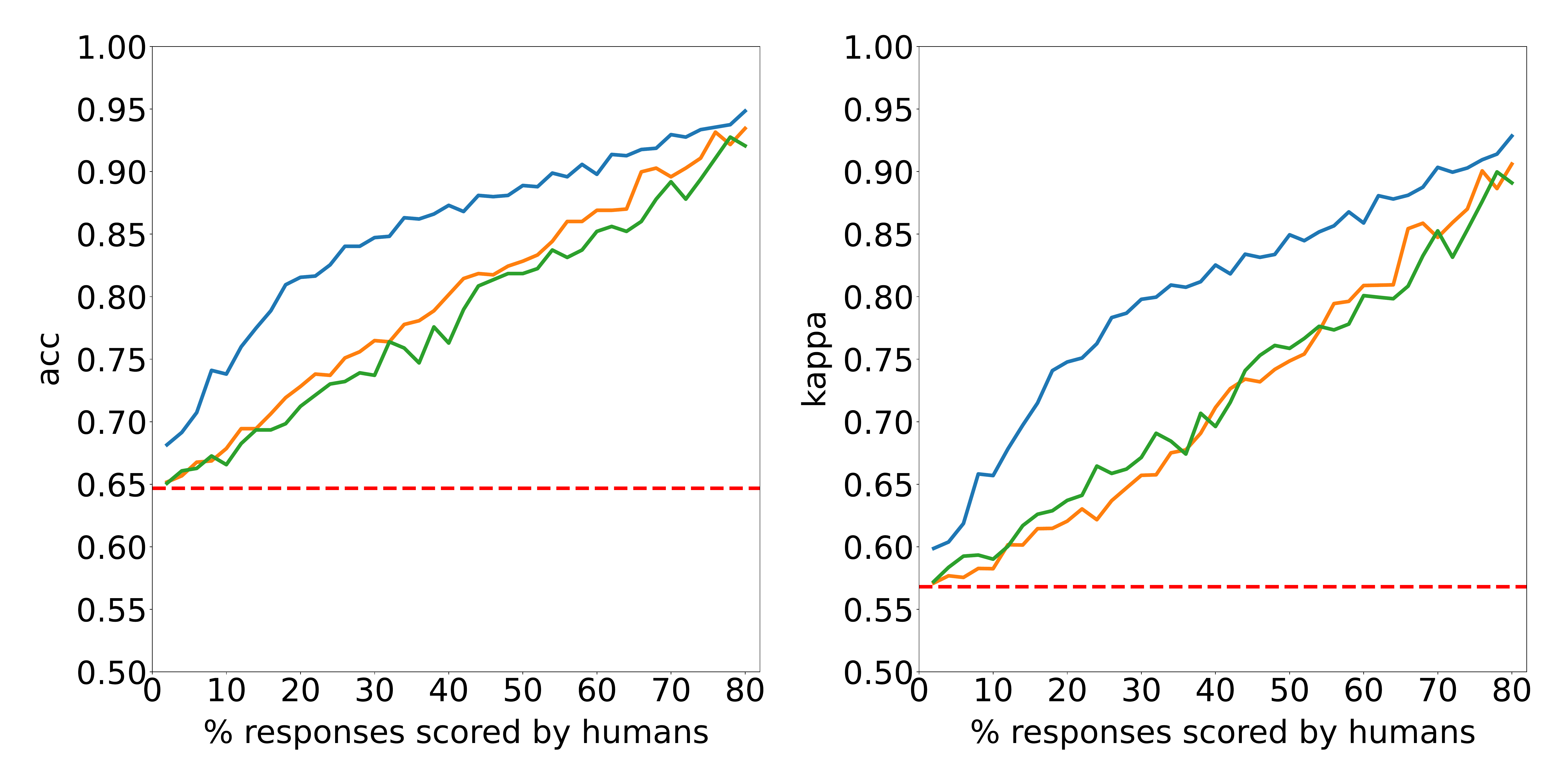}
    \caption{LSTM-Attention-Two Stage Model}
    \end{subfigure}
    \begin{subfigure}[b]{0.4\textwidth}
    \includegraphics[width=\columnwidth]{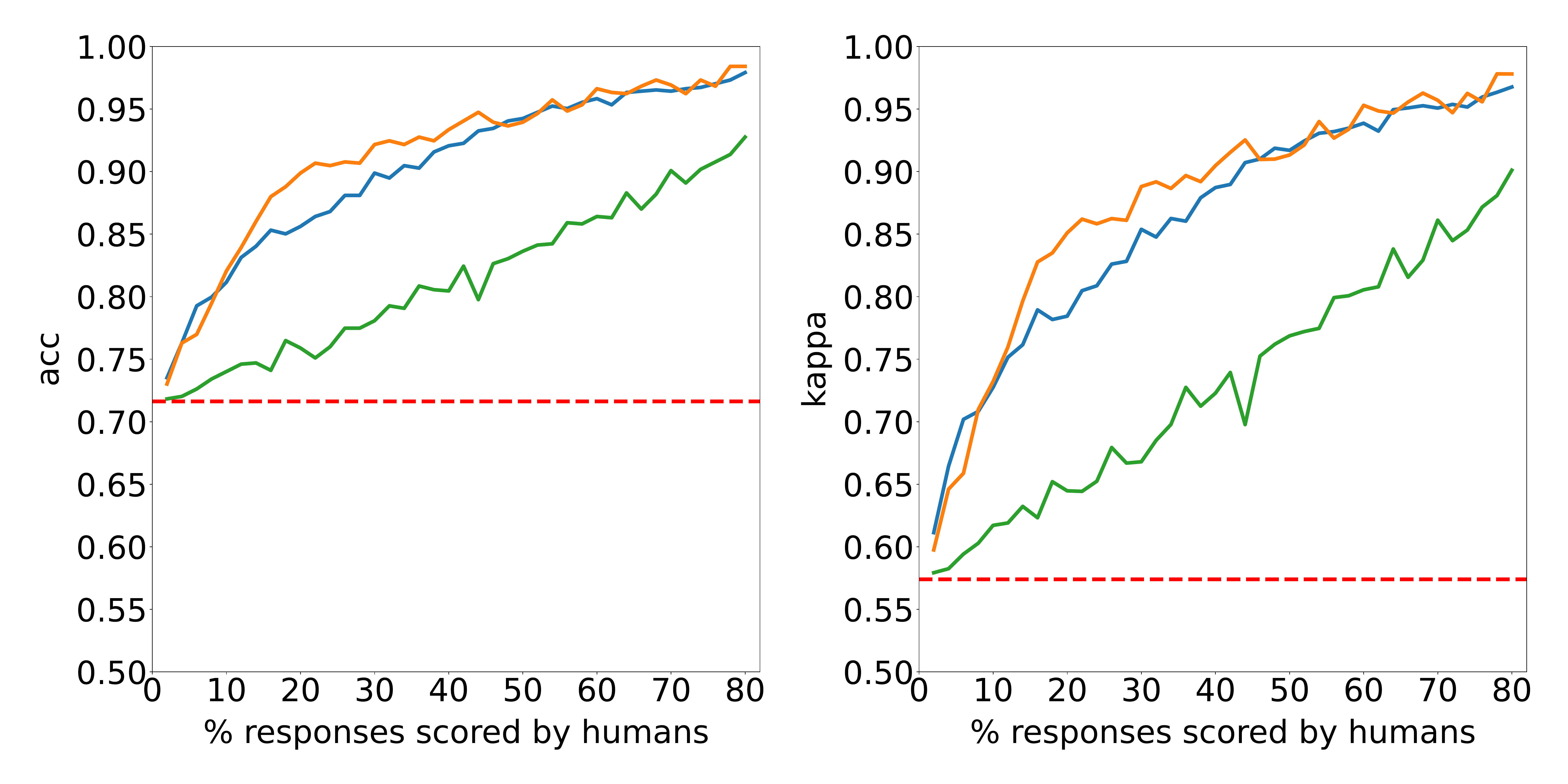}
    \caption{Pseudo Model with 0.75 accuracy (local)}
    \end{subfigure}%
    \begin{subfigure}[b]{0.4\textwidth}
    \centering
    \includegraphics[width=0.47\columnwidth]{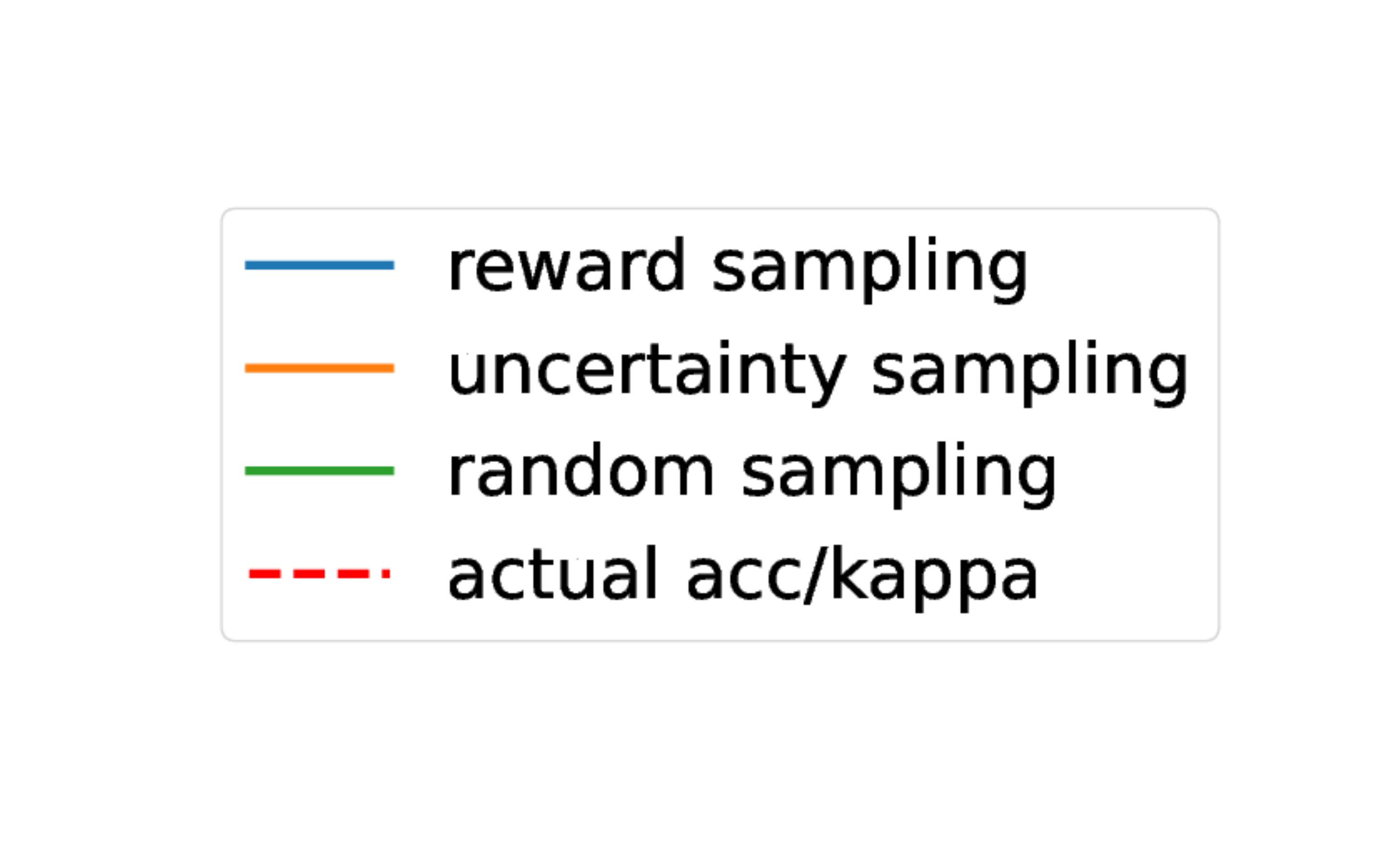}
    \caption{Legend}
    \end{subfigure}
    \caption{In each model, we show the change in accuracy (\textit{left}) and quadratic weighted kappa (QWK) (\textit{right}) after sampling with the sample size (human budget) shown on the x-axis. As can be seen, reward sampling outperforms both uncertainty sampling and random sampling baseline in each model.}
    \label{merged_model_acc}
\end{figure*}

\section{Experiments}

\subsection{Dataset}

To evaluate our method, we make use of data collected by Second Language Testing Inc. (SLTI) while conducting the Simulated Oral Proficiency (SOPI) Exam. The SOPI exam has been used since 1992 and studied extensively \cite{stansfield1992development, stansfield1996test}. SOPI is used for interviews, university admissions, skill development and as a test in several online courses \cite{slti}. SOPI offers psychometric advantages in terms of reliability and validity, particularly in standardized testing situations. The candidates in the dataset are primarily Filipino high school graduates and
above. A test-taker is presented with six prompts on their computer and their responses for each individual item are recorded. The prompts and the rubrics for evaluation follow the Central European Framework of Reference for Languages (CEFR) \cite{broeder2008language} guidelines. The prompts difficulty varies from B1 to C1. A candidate receives both a prompt-level score and a global score calculated from the individual prompt-level scores. The SOPI dataset has eight question papers (forms) containing six prompts each, and each form was attempted by 7200 speakers on an average. Many other works have used the SLTI dataset for tasks including automated scoring and coherence modeling \cite{grover2020multi, patil2020towards, stansfield2008testing, singla2021speaker}.

\subsection{Experimental Setup}

To demonstrate that the sampling methods described are model agnostic, we conduct experiments using multiple models of varying accuracy.
We leverage speech scoring models from \citet{singla2021speaker}, making use of state-of-the-art models such as BERT and Bi-directional LSTMs, both baseline versions and conditioned on speaker information. In addition, we also run experiments on a pseudo model, described as follows. For a given accuracy, a pseudo model's predictions are generated by randomly changing $100-acc\%$ of ground truth labels. For \textit{e.g.}, the prediction of a pseudo model with 65\% accuracy is the ground truth with 35\% of labels randomly changed. Predictions, and hence accuracy, are generated at the \textit{local} level, for each response whereas we are concerned about the metrics at the global level, which is typically lesser. The dataset is split into train and test sets, with the additional constraint that this split be done such that all responses of one candidate are contained in a set, and not split between the train and test sets. For our experiments, since we do not have a precomputed human-machine agreement matrix, we compute it using the training set and hold out the test set for verifying our proposed system. In addition, the aggregate dataset must also be calculated from each candidate's individual responses, calculating the candidate's global score from each of their responses.

The experiments were conducted with sample sizes upto 80\% of the dataset to observe the effect of sample size on the improvement in accuracy with respect to each sampling method. Records are sampled from the test set according to Reward Sampling, along with Random (uniform) Sampling and Uncertainty (importance) Sampling, our baselines. We replace the predictions of records in the sample with those of the ground truth, following which we recompute the aggregate dataset. This dataset is used to calculate the \textbf{system level metrics}, not just the model, but the combination of the model and the human in the loop. In estimating metrics, a secondary sample was taken. Empirically, we observed that a sample size of 200 was sufficient for stable estimations of a 95\% confidence interval. We report our estimation on Reward Sampling when considering 80\% of the dataset as human budget {\it i.e.} the most performant configuration. 

\begin{figure}[t]
\centering
\includegraphics[width=\columnwidth]{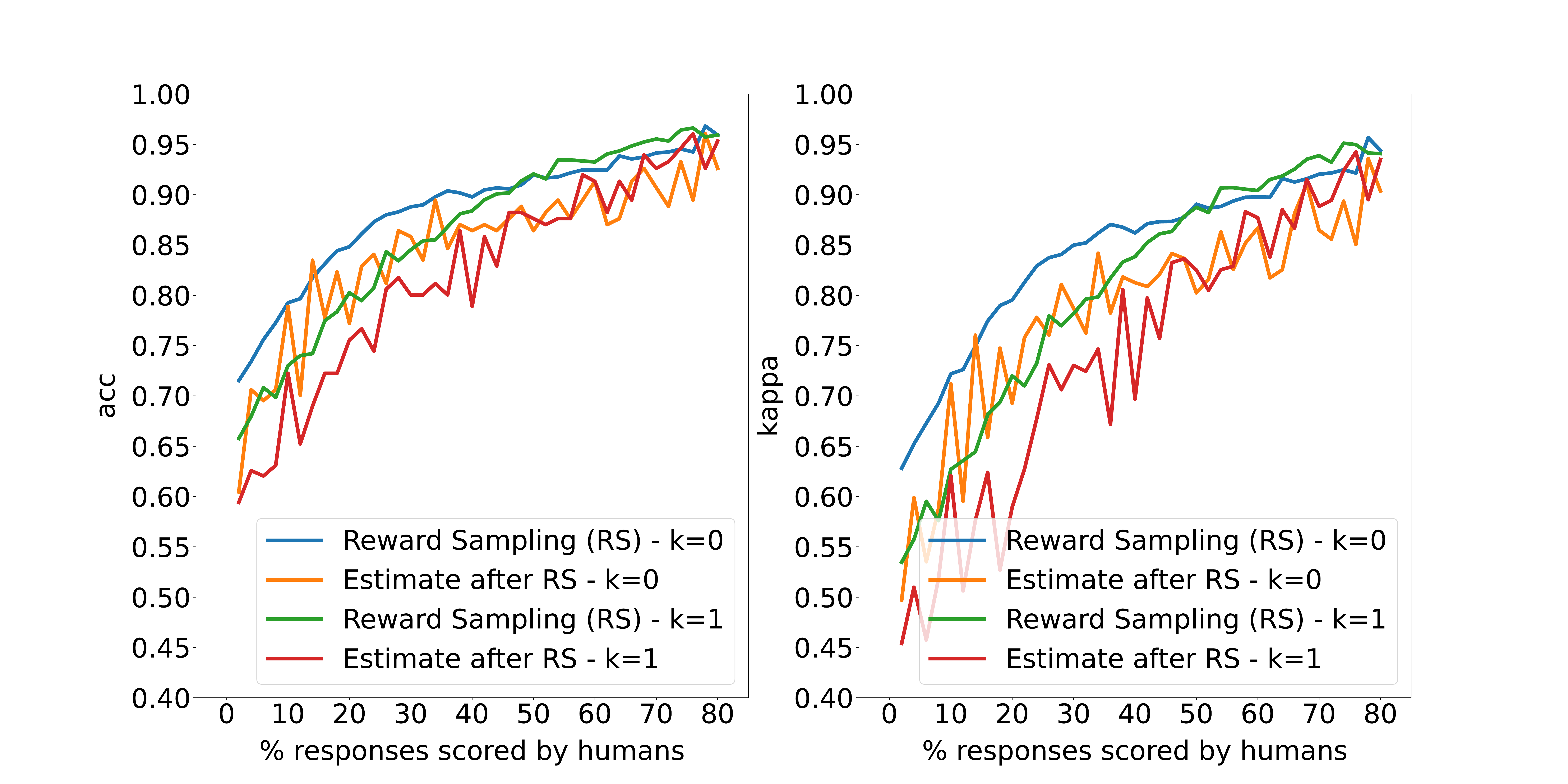}
\caption{The metrics of two models [0 - BERT-TwoStage, 1 - LSTM-Baseline] have been presented. Results for other models are similar and are not shown for visual clarity.}
\label{fig:model_estim}
\end{figure}

\section{Results}

\subsection{Improving Reliability}

Table~\ref{result_improving_metrics} presents the results of the experiments conducted across configurations of models, sampling methods, and human budget. Entries in \textbf{bold} indicate the best performing configuration, which is identical across nearly all models (Reward Sampling with the maximum sample size, 80\% of the dataset). The models considered are BERT (baseline), BERT (two stage speaker conditioning), BD-LSTM with Attention (baseline), BD-LSTM with Attention (two stage speaker conditioning) and a pseudo model with $accuracy = 0.75$ at the local level. For all models, the dataset is aggregated, following which accuracy and QWK are calculated, giving us the values in the \textit{Model metrics} column.

Fig~\ref{merged_model_acc} shows the change in both accuracy and QWK at the \textbf{global} level for various models. The changes are measured for increasing human budget {\it i.e.} percent of responses available to be scored by humans upto 80\% of the dataset. For illustration, we consider the BERT-Baseline model. Firstly, we observe that random sampling shows minimal improvement (\textbf{3\%}) over the actual accuracy when sampling 10\% of the dataset to be scored by human raters. This is due to the model's accuracy (66\% of all samples would have been predicted correctly anyway) and the remaining gains are further minimized by the aggregation process. Reward Sampling, on the other hand, shows an 8\% gain in accuracy, more than twice the gains achieved by random sampling. Interestingly, Uncertainty Sampling shows similar gains to random sampling in all models except the pseudo model, where its performance is more in line with reward sampling. The predictions of the pseudo model are randomly generated, hence local gains translate well to global gains. This difference is likely the reason for the large gap in performance when considering uncertainty sampling on pseudo and real models. 

Reward Sampling, where the reward that is gained by having a record rated by a human is also a sampling factor, shows significant gains across models as shown in Fig~\ref{merged_model_acc}. We note that the gains provided by Reward Sampling decline compared to the baseline sampling methods with increasing sample sizes. Initially, Reward Sampling outperforms the other sampling methods with large gains, upto a sample size of $\sim$30\%. Beyond this mark, the gains are no longer as significant and the other methods slowly catch up. This trend holds across all models, indicating that Reward Sampling shows maximal gains over baselines when sampling less than half of the dataset for human scoring.

\subsection{Estimation with Guarantees}

Fig~\ref{fig:model_estim} is a plot visualizing the metrics of the models when utilizing reward sampling and an estimate of the same. The sample size used for estimation remains constant and it is only the sample used for reward sampling that changes. After reward sampling, a sample based on the confidence distribution (\S\ref{sec:Estimation with Guarantees}) is drawn, and the 95\% confidence interval for both accuracy and kappa is calculated. The lower bound is taken to provide a statistical guarantee that accuracy/QWK will only fall below the estimated values 5\% of all runs.

\section{Conclusion}
Automatic Scoring (AS) helps assess the language competency of candidates with accuracy matching that of a human grader, but faster, with greater consistency and at a fraction of the cost. Existing systems either rely on double scoring, effectively scoring each sample by both human and AS system, or solely by an AS system. Although double scoring is more reliable, it is considerably more expensive. We develop novel, sample-efficient algorithms to target the spectrum of possible solutions in the middle of both extremes. We show that by using a relatively small human budget, we can improve and estimate performance with guarantees, thus increasing the reliability and trustworthiness of the system. We implement and evaluate our algorithms on real exam data, showing that they outperform naive baselines in all settings evaluated. These results indicate the promise of probabilistic algorithms to improve and estimate automatic scoring reliability with statistical guarantees. 


As part of future research, we plan to work on even more sample efficient algorithms and incorporating trait scoring while sampling. Another possible research avenue where we can apply our algorithms is in test design. While right now test design involves linguistic validity assessment studies, it does not take into account the reliability of the final test built. Reliability of a test could be incorporated as another constraint easily through our modelling paradigm.

\bibliography{references}

\end{document}